\documentclass[twocolumn]{el-author}
\usepackage{amsmath, amssymb}
\usepackage{multirow}
\usepackage{dsfont}
\usepackage{float}
\usepackage{subfigure}

\newcommand{\ua}{\uparrow}
\newcommand{\nc}{\newcommand}
\nc{\da}{\downarrow} \nc{\hc}{\hat{c}} \nc{\hS}{\hat{S}}
\nc{\bra}{\langle} \nc{\ket}{\rangle} \nc{\eq}{equation (\ref}
\nc{\h}{\hat} \nc{\hT}{\h{T}}\nc{\be}{\begin{eqnarray}}
\nc{\ee}{\end{eqnarray}}\nc{\rd}{\textrm{d}}\nc{\e}{eqnarray}\nc{\hR}{\hat{R}}\nc{\Tr}{\mathrm{Tr}}
\nc{\tS}{\tilde{S}}\nc{\tr}{\mathrm{tr}}\nc{\8}{\infty}\nc{\lgs}{\bra\ua,\phi|}\nc{\rgs}{|\ua,\phi\ket}
\nc{\hU}{\hat{U}}\nc{\lfs}{\bra\phi|}\nc{\rfs}{|\phi\ket}\nc{\hZ}{\hat{Z}}\nc{\hd}{\hat{d}}\nc{\mD}{\mathcal{D}}
\nc{\bd}{\bar{d}}\nc{\bc}{\bar{c}}\nc{\mc}{\mathcal}\nc{\ea}{eqnarray}\nc{\mG}{\mathcal{G}}\nc{\bce}{\begin{center}}
\nc{\ece}{\end{center}}
\date{31th January 2013}

\begin{document}

\title{Inductive Sparse Subspace Clustering}
\author{X. Peng, L. Zhang and Z. Yi}

\abstract{
    Sparse Subspace Clustering (SSC) has achieved state-of-the-art clustering quality by performing spectral clustering over a $\ell^{1}$-norm based similarity graph. However, SSC is a transductive method which does not handle with the data not used to construct the graph (out-of-sample data). For each new datum, SSC requires solving $n$ optimization problems in $O(n)$ variables for performing the algorithm over the whole data set, where $n$ is the number of data points. Therefore, it is inefficient to apply SSC in fast online clustering and scalable graphing. In this letter, we propose an inductive spectral clustering algorithm, called inductive Sparse Subspace Clustering (iSSC), which makes SSC feasible to cluster out-of-sample data. iSSC adopts the assumption that high-dimensional data actually lie on the low-dimensional manifold such that out-of-sample data could be grouped in the embedding space learned from in-sample data. Experimental results show that iSSC is promising in clustering out-of-sample data.
}

\maketitle

\section{Introduction}

Spectral clustering is one of the most popular subspace clustering algorithms, which aims to find a cluster membership of data points and the corresponding low-dimensional representation by utilizing the spectrum of a Laplacian matrix. The entries in the Laplacian matrix depict the similarity among data points. Thus, the construction of similarity graph lies on the heart of spectral clustering. In a similarity graph, the vertex denotes a data point and the connection weight between two points represents their similarity.

Recently, Elhamifar and Vidal~\cite{Elhamifar2012} constructed a similarity graph by using $\ell^1$-minimization based coefficient and performed spectral clustering over the graph, named Sparse Subspace Clustering (SSC). It automatically selects the nearby points for each datum by utilizing the principle of sparsity without pre-determination of the size of neighborhood. SSC has achieved impressive performance in images clustering and motion segmentation. However, it requires solving $n$ optimization problems over $n$ data points and calculating the eigenvectors of a $n\times n$ matrix, resulting in a very high computational complexity. In general, the time complexity of SSC is proportion to the cubic of data size. Thus, any medium-sized data set will bring up the scalability issues with SSC. In addition, SSC is a transductive algorithm which does not handle with the data not used to construct the graph (out-of-sample data). For each new datum, SSC needs performing the algorithm over the whole data set, which makes SSC inefficient to fast online clustering and scalable grouping.

To address the scalability issue and the out-of-sample problem in SSC, we propose an inductive clustering algorithm which is called inductive Sparse Subspace Clustering algorithm (iSSC). Out motivation derives from a widely-accepted assumption in manifold learning that the high-dimensional data actually lie on the low-dimensional manifold. Therefore, we could obtain the cluster membership of out-of-sample data by assigning them to the nearest cluster in the embedding space learned from well-sampled in-sample data. In other words, we resolve the out-of-sample problem in SSC by using subspace learning method. On the other hand, for large scale data set, we randomly split it into two parts, in-sample data and out-of-sample data, such that scalability issue could be addressed as an out-of-sample problem.

Except in some specified cases, \textbf{lower-case bold letters} represent column vectors and \textbf{upper-case bold ones} represent matrices. $\mathbf{A}^T$ denotes the transpose of the matrix $\mathbf{A}$ whose pseudo-inverse is $\mathbf{A}^{-1}$, and $\mathbf{I}$ is reserved for identity matrix.

\section{Inductive Sparse Subspace Clustering Algorithm}

\begin{figure}[h]
\centering{
\subfigure []{\label{fig:1.a}\centering\includegraphics[width=0.16\textwidth]{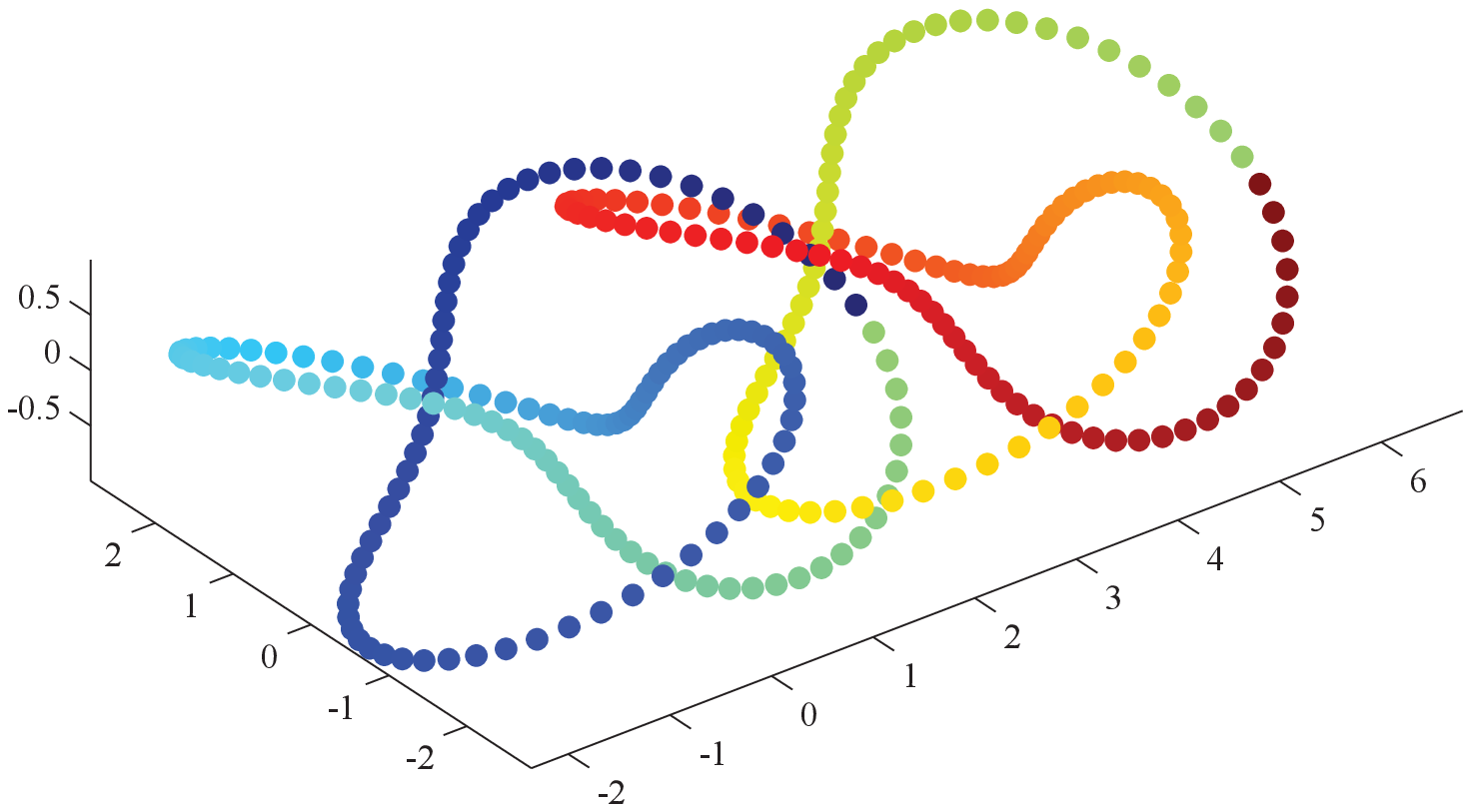}}
\subfigure []{\label{fig:1.b}\centering\includegraphics[width=0.16\textwidth]{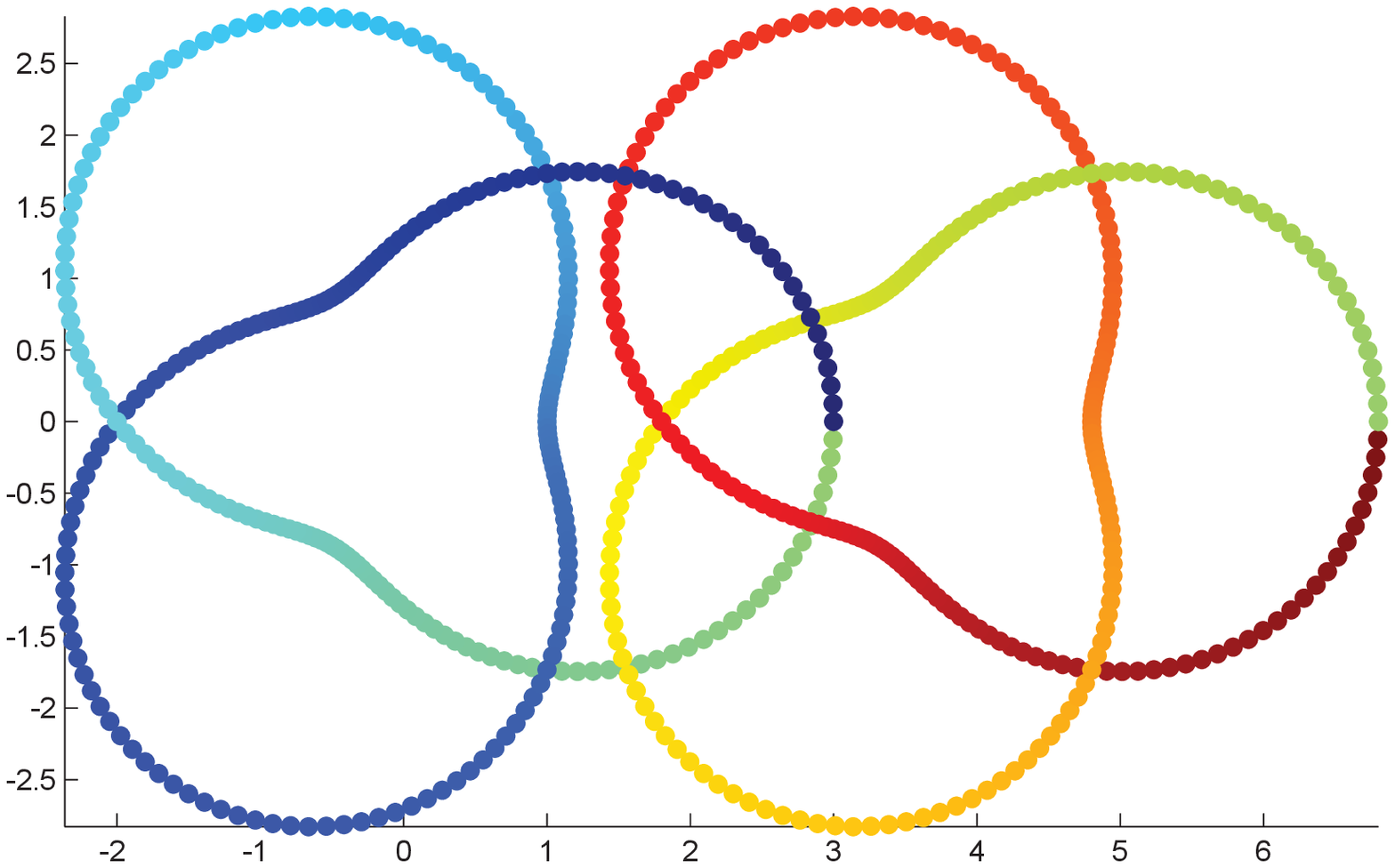}}
\subfigure []{\label{fig:1.c}\centering\includegraphics[width=0.16\textwidth]{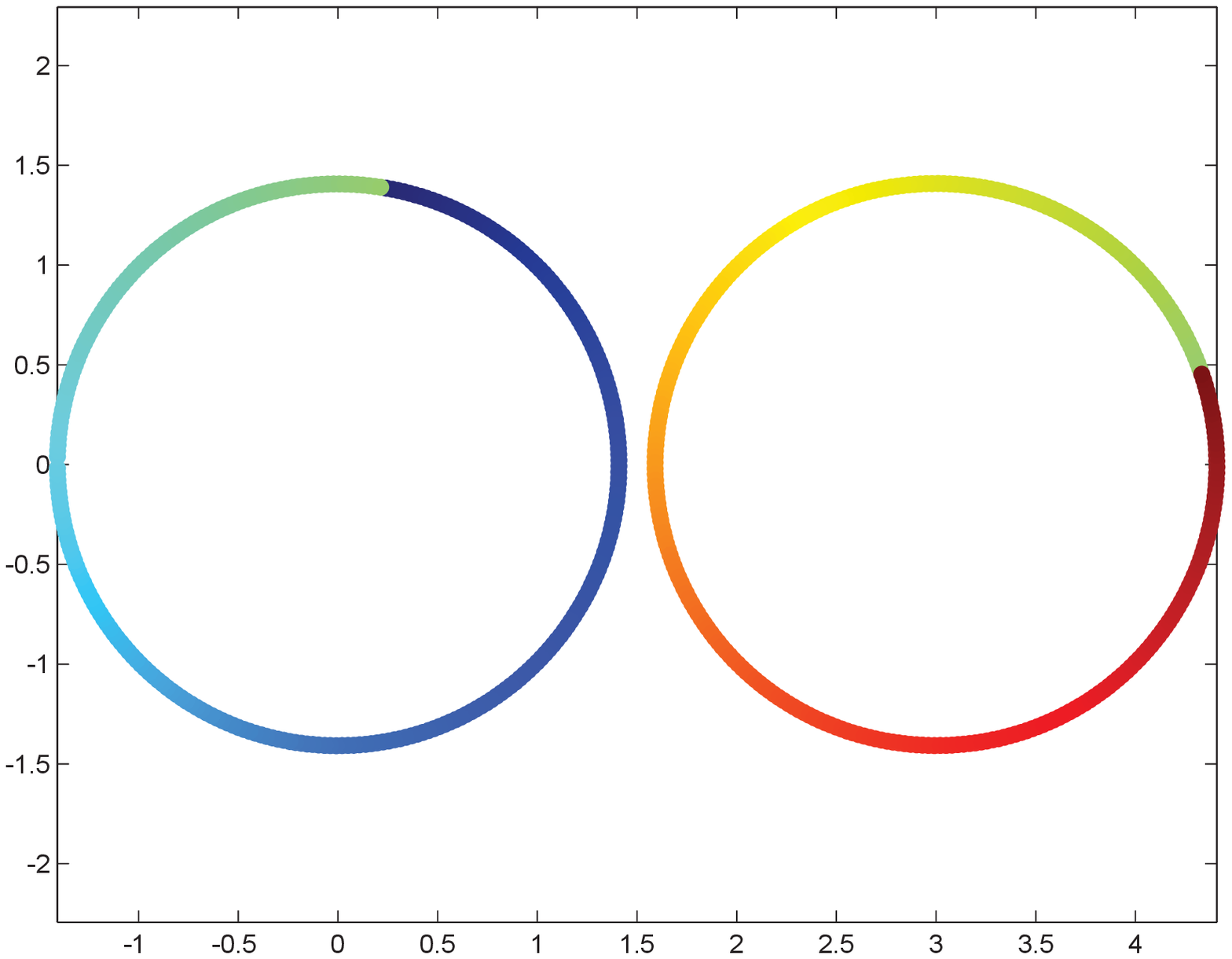}}
}
\caption{A key observation. (a) some data points sampled from two 2-dimensional manifolds (trefoil-knots) which are embedded into a 3-dimensional space; (b) a plan view of the sampled data; (c) the embedding of the sampled data. It is easy to find that out-of-sample data points could be easily grouped into the correct cluster after they were projected into the embedding space.}
\label{fig:1}
\end{figure}

The basic idea of our approach is that: Suppose two data sets $\mathbf{Y}\in \mathds{R}^{m\times p}$ (in-sample data) and $\mathbf{X}\in \mathds{R}^{m\times n}$ (out-of-sample data) are drawn from multiple underlying manifolds of which each corresponds to a subspace. Provided $\mathbf{Y}$ is sufficient such that the manifolds are well-sampled, we expect to learn an embedding space with $\mathbf{Y}$ and group $\mathbf{X}$ in the embedding space since it is more compact and discriminative than the original space (See Fig.~\ref{fig:1}).

We make SSC feasible to cluster out-of-sample data in "subspace clustering, subspace learning and extension" manner. The first two steps are offline processes which only involve in-sample data, and the last one groups the out-of-sample data in online way.

To obtain the cluster membership of in-sample data $\mathbf{Y}$, iSSC firstly constructs a similarity graph by minimizing the following objective function,
\begin{align}
\label{equ:1} \mathrm{min}\hspace{1mm}\|\mathbf{c}_i\|_1
\hspace{8mm} \mathrm{s.t.} \hspace{2mm}
\|\mathbf{y}_i-\mathbf{Y}_i \mathbf{c}_{i}\|_2<\delta,
\end{align}
where $\mathbf{c}_i\in\mathds{R}^{p}$ is the sparse representation of the data point $\mathbf{y}_i\in\mathds{R}^m$ over the dictionary $\mathbf{Y}_i \triangleq [\mathbf{y}_1 \ldots \mathbf{y}_{i-1}\ \mathbf{0}\ \mathbf{y}_{i+1} \ldots \mathbf{y}_p]$, and $\delta\ge 0$ is the error tolerance.

After getting the coefficients of $\mathbf{Y}$, iSSC performs normalized spectral clustering over the sparse coefficients to get the cluster membership of $\mathbf{Y}$ as SSC does. However, SSC could not efficiently cope with out-of-sample data. Motivated by the assumption in manifold learning, we aim to group out-of-sample data in the embedding space. In this letter, following the embedding program of Neighborhood Preserving Embedding algorithm (NPE)~\cite{He2005}, we perform subspace learning to compute the projection matrix $\mathbf{W}$ via
\begin{align}
\label{equ:2}
    \mathop{\mathrm{min}}_{\mathbf{W}}
        \left\| \mathbf{W} ^T\mathbf{Y} - \mathbf{W}^T\mathbf{Y} \mathbf{C} \right\|_2^2, \mathrm{\hspace{2mm}
        s.t. \hspace{1mm} \mathbf{W}^T\mathbf{Y}\mathbf{Y}^T\mathbf{W}=\mathbf{I}},
\end{align}
where $\mathbf{C}\in \mathds{R}^{p\times p}$ is the collection of the sparse representation of $\mathbf{Y}$ produced by (\ref{equ:1}), and the constraint term aims at the scale-invariance.

The solution of (\ref{equ:2}) is given by the maximum eigenvalue solution to the following generalized eigenvector problem:
\begin{align}
\label{equ:3}
    \mathbf{W}^{T}\mathbf{Y} (\mathbf{C} + \mathbf{C}^T - \mathbf{C}^T \mathbf{C}) \mathbf{Y}^T \mathbf{W}
    =\lambda \mathbf{W}^{T}\mathbf{Y} \mathbf{Y}^T \mathbf{W}
\end{align}

Once the optimal $\mathbf{W}$ is achieved, iSSC transforms out-of-sample data $\mathbf{X}$ in the embedding space via $\mathbf{W}^T\mathbf{X}$, and then assigns $\mathbf{X}$ to the nearest cluster in the space.

The steps of iSSC can be summarized as follows:
\begin{enumerate}
  \item For in-sample data $\mathbf{Y}$, calculate the sparse representation coefficients $\mathbf{C}$ via solving
        \begin{align*}
         \label{equ:4} \mathrm{min}\hspace{1mm}\|\mathbf{c}_i\|_1
            \hspace{8mm} \mathrm{s.t.} \hspace{2mm}
            \|\mathbf{y}_i-\mathbf{Y}_i \mathbf{c}_{i}\|_2<\delta.
        \end{align*}
  \item Construct a Laplacian matrix $\mathbf{L}=\mathbf{S}^{-\frac{1}{2}} \mathbf{A} \mathbf{S}^{-\frac{1}{2}}$ by using the affinity matrix $\mathbf{A}$, where $\mathbf{S}=\mathrm{diag}\{s_{i}\}$ with $s_{i}=\sum_{j=1}^{p}a_{ij}$, $a_{ij}$ is an entry of $\mathbf{A}$ and $\mathbf{A}=|\mathbf{C}|+|\mathbf{C}|^T$.
  \item Obtain the eigenvector matrix $\mathbf{V}\in \mathds{R}^{p\times k}$ which consists of the first $k$ normalized eigenvectors of $\mathbf{L}$ corresponding to its $k$ smallest eigenvalues.
  \item Get the segmentations of $\mathbf{Y}$ by performing k-means clustering algorithm on the rows of $\mathbf{V}$.
  \item Suppose the desired dimensionality of embedding space is $d$, the projection matrix $\mathbf{W}\in \mathds{R}^{m\times d}$ is given by the eigenvectors with $d$ largest eigenvalues of the following eigenvector problem:
       \begin{equation*}
        \label{equ:5}
        \mathbf{M} \mathbf{Z}^T = \lambda \mathbf{Z}^T,
       \end{equation*}
       where $\mathbf{M} = \mathbf{C} + \mathbf{C}^T - \mathbf{C}^T \mathbf{C}$ and $\mathbf{Z}=\mathbf{W}^T \mathbf{Y}$.
  \item Project out-of-sample data $\mathbf{X}$ into the $d$-dimensional space via $\mathbf{W}^T\mathbf{X}$.
  \item Search the nearest neighbor of $\mathbf{X}$ from $\mathbf{Y}$ in the embedding space, and assign $\mathbf{X}$ to the cluster that the neighbor belongs to.
\end{enumerate}


\section{Computational Complexity Analysis}


Suppose in-sample data $\mathbf{Y}\in \mathbf{R}^{m\times p}$ drawn from $k$ subspaces, we need $O(t_1 mp^3+t_2pk^2)$ to perform SSC over $\mathbf{Y}$, where $t_1$ and $t_2$ are the numbers of iteration of Homotopy optimizer~\cite{Yang2010} and k-means clustering algorithm, respectively. Moreover, we need $O(p^3)$ to compute the projection matrix $\mathbf{W}^T$. To group out-of-sample datum $\mathbf{X}\in \mathbf{R}^{m\times n}$, we need $O(dmn)$ to obtain its $d$-dimensional representation and $O(dpn)$ to search the nearest neighbor of $\mathbf{X}$ from $\mathbf{Y}$ in the embedding space.

Putting everything together, the computational complexity of iSSC is $O(t_1 mp^3+t_2pk^2+dpn)$ owing to $d<m$ and $m<p$, where $m<p$ derives from the conditions of compressive sensing theory. Clearly, under the same conditions, iSSC is more efficient than SSC whose time complexity is about $O(t_1 mn^3+t_2nk^2)$ .

\section{Baselines and Evaluation Metrics}
We presented the experimental results of our approach over three real-world data sets, i.e., Extended Yale Database B (ExYaleB)~\cite{Georghiades2001}, Pendigit\footnote{http://archive.ics.uci.edu/ml/datasets.html} and USPS\footnote{http://www.cs.nyu.edu/~roweis/data.html}. ExYaleB contains 2414 facial images of 38 subjects. We cropped the images from $192\times 168$ to $48\times 42$ and extracted 114 features by using PCA to retain 98\% energy of the cropped data. Pendigit and USPS are two handwritten digital data sets distributed over 10 classes, where Pendigit contains 10992 samples with 54 features and USPS consists of 11000 samples with 256 dimensionality.

We compared iSSC with three state-of-the-art inductive clustering algorithms, i.e., Nystr\"{o}m based spectral clustering~\cite{Chen2011}, Spectral Embedding Clustering (SEC)~\cite{Nie2011} and Approximate Kernel K-means (AKK)~\cite{Chitta2011}. Note that, Nystr\"{o}m based spectral clustering and SEC have two variants, respectively. We denote the variants as Nystr\"{o}m, Nystr\"{o}m\_Orth, SEC\_K and SEC\_R. The approximate affinity matrix of Nystr\"{o}m is non-orthogonal, while that of Nystr\"{o}m-Orth is column-orthogonal. SEC\_K performs k-means to get the clustering results and SEC\_R adopts spectral rotation method to obtain the final cluster assignment matrix. Furthermore, we also reported the results of k-means clustering as a baseline. The MATLAB code of iSSC can be downloaded at https://www.dropbox.com/s/ju6y9qe7w8lcdyp/CodeAndData.zip.

We adopted two widely-used metrics, Accuracy and Normalized Mutual Information (NMI), to measure the clustering quality of the tested methods. The value of Accuracy or NMI is 1 indicates that the predicted clustering membership is totally matching with the ground truth whereas 0 indicates totally mismatch.

In all experiments, the tuned parameters for the algorithms were applied to achieve their best Accuracy. Specifically, iSSC adopted Homotopy optimizer~\cite{Yang2010} to solve $\ell^1$-minimization problem. The optimizer needs two user-specified parameters, sparsity parameter $\lambda$ and error tolerance parameter $\delta$. We found a good value combination by setting $\lambda=(10^{-7},10^{-6},10^{-5})$ and $\delta=(10^{-3},10^{-2},10^{-1})$. Moreover, iSSC groups out-of-sample data in an low-dimensional space which preserves 98\% energy of the embedding space learned from in-sample data. For the other competing methods, we set the value range for different parameters by following the configurations in~\cite{Chen2011,Nie2011,Chitta2011}.


\section{Results}

To examine the effectiveness of the algorithms, we randomly selected a half of images (1212) from ExYaleB as in-sample data and used the remaining samples as out-of-sample data. In the similar way, we formed two data sets by choosing 1000 samples from Pendigit and USPS as in-sample data and used the rest as out-of-sample data, respectively.

\begin{table}[h]
\processtable{Performance comparisons in different algorithms over ExYaleB.}
{\begin{tabular}{|c|c|c|c|}
\hline
Algorithms & Accuracy & NMI & Time(s)\\
\hline
iSSC (1e-6, 1e-3) & \textbf{59.69}\% & \textbf{62.77}\% & 24.88 \\
\hline
Nystr\"{o}m (12) & 25.72\% & 46.57\% & \textbf{9.33} \\
\hline
Nystr\"{o}m\_Orth (2) & 21.71\% & 41.74\% & 58.87 \\
\hline
SEC\_K (1e+12, 5, 1) & 11.02\% & 11.09\% & 34.91\\
\hline
SEC\_R (1e+9, 4, 1) & 5.97\% & 4.31\% & 19.96\\
\hline
AKK (0.4) & 8.00\% & 9.01\% & 9.94 \\
\hline
k-means & 9.03\% & 11.20\% & 37.05 \\
\hline
\end{tabular}\label{tab:1}}{}
\end{table}

\begin{table}[h]
\processtable{Performance comparisons in different algorithms over Pendigit.}
{\begin{tabular}{|c|c|c|c|}
\hline
Algorithms & Accuracy & NMI & Time(s)\\
\hline
iSSC (1e-6, 0.1) & \textbf{84.94}\% & \textbf{71.17}\% & 28.03\\
\hline
Nystr\"{o}m (0.8) & 76.09\% & 68.10\% & \textbf{8.54}\\
\hline
Nystr\"{o}m\_Orth (6) & 75.72\% & 67.04\% & 39.24\\
\hline
SEC\_K (1e-6, 4, 1) & 76.91\% & 63.84\% & 24.51\\
\hline
SEC\_R (1e-9, 1, 1) & 11.01\% & 1.18\% & 18.14\\
\hline
AKK (0.9) & 77.22\% & 69.48\% & 9.80\\
\hline
k-means & 77.05\% & 69.21\% & 30.21 \\
\hline
\end{tabular}\label{tab:2}}{}
\end{table}

\begin{table}[h]
\processtable{Performance comparisons in different algorithms over USPS.}
{\begin{tabular}{|c|c|c|c|}
\hline
Algorithms & Accuracy & NMI & Time(s)\\
\hline
iSSC (1e-7, 0.01) & \textbf{52.93}\% & \textbf{52.90}\% & 41.52 \\
\hline
Nystr\"{o}m (14) & 47.66\% & 44.42\% & \textbf{15.91} \\
\hline
Nystr\"{o}m\_Orth (0.5) & 50.70\% & 44.60\% & 183.37 \\
\hline
SEC\_K (1e-9, 3, 1) & 47.63\% & 42.28\% & 43.38\\
\hline
SEC\_R (1e-6, 4, 1) & 11.70\% & 1.44\% & 19.78\\
\hline
AKK (0.3) & 48.49\% & 46.79\% & 16.81 \\
\hline
k-means & 46.54\% & 45.61\% & 250.82 \\
\hline
\end{tabular}\label{tab:3}}{}
\end{table}

Tables~\ref{tab:1}-\ref{tab:3} report the clustering quality and the time costs of the tested algorithms over the data sets. In the parenthesis, we also show the tuned parameters when the best Accuracy was achieved. From the results, we have the following observations:
\begin{itemize}
  \item In all the tests, iSSC demonstrates an elegant balance between running time and clustering quality. Although iSSC is not the fastest algorithm, it outperforms the other tested methods with considerable performance margins in Accuracy and NMI. For example, iSSC achieved 33.97\% gains in Accuracy and 16.20\% gains in NMI over the second best algorithm when ExYaleB database was used to test.
  \item The accelerating kernel-based method (AKK) is more competitive when it was applied to cluster handwritten digital data but facial data. Moreover, AKK performed very close to k-means algorithm, which is consistent with the results in~\cite{Chitta2011}.
\end{itemize}


\section{Conclusion}

In this letter, we have presented an inductive spectral clustering algorithm, called inductive Sparse Subspace Clustering (iSSC). The algorithm, which is an out-of-sample extension of Sparse Subspace Clustering algorithm (SSC)~\cite{Elhamifar2012}, scales linearly with the problem size such that it could be applied to fast online learning. Experimental results with facial image and digital image clustering indicate the effectiveness of iSSC comparing with the state-of-the-art approaches.

\vskip3pt
\ack{This work has been supported by ...}

\end{document}